\documentclass[12pt]{article}
\usepackage{amsmath,amsfonts}
\usepackage{algorithmic}
\usepackage{algorithm}
\usepackage{array}
\usepackage{subcaption}
\usepackage[caption=false,font=normalsize,labelfont=sf,textfont=sf]{subfig}
\usepackage{textcomp}
\usepackage{stfloats}
\usepackage{url}
\usepackage{verbatim}
\usepackage{graphicx}
\usepackage{cite}
\usepackage{multirow}
\usepackage{tabularx}
\usepackage{makecell}
\usepackage[left=1in, right=1in, top=0.5in, bottom=1in]{geometry} 
\usepackage{authblk}

\begin{document}

\title{Machine Learning-Driven Student Performance Prediction for Enhancing Tiered Instruction}

\author[1]{Yawen Chen}
\author[1]{Jiande Sun}
\author[1]{Jinhui Wang}
\author[23]{Liang Zhao}
\author[1]{Xinmin Song}
\author[1]{Linbo Zhai}

\affil[1]{School of Information Science and Engineering, Shandong Normal University}
\affil[2]{Shandong Provincial Institute of Education Sciences}
\affil[3]{Faculty of Education, Shandong Normal University}

\date{}
\maketitle

\begin{abstract}

Student performance prediction is one of the most important subjects in educational data mining. As a modern technology, machine learning offers powerful capabilities in feature extraction and data modeling, providing essential support for diverse application scenarios, as evidenced by recent studies confirming its effectiveness in educational data mining. However, despite extensive prediction experiments, machine learning methods have not been effectively integrated into practical teaching strategies, hindering their application in modern education. In addition, massive features as input variables for machine learning algorithms often leads to information redundancy, which can negatively impact prediction accuracy. Therefore, how to effectively use machine learning methods to predict student performance and integrate the prediction results with actual teaching scenarios is a worthy research subject. To this end, this study integrates the results of machine learning-based student performance prediction with tiered instruction, aiming to enhance student outcomes in target course, which is significant for the application of educational data mining in contemporary teaching scenarios. Specifically, we collect original educational data and perform feature selection to reduce information redundancy. Then, the performance of five representative machine learning methods is analyzed and discussed with Random Forest showing the best performance. Furthermore, based on the results of the classification of students, tiered instruction is applied accordingly, and different teaching objectives and contents are set for all levels of students. The comparison of teaching outcomes between the control and experimental classes, along with the analysis of questionnaire results, demonstrates the effectiveness of the proposed framework.

\vspace{0.5cm}


\end{abstract}

\section{Introduction}

The advancement of modern educational technology, significantly boosted by the application of data mining techniques, has transformed traditional education to meet the evolving needs of teachers and learners. Educational Data Mining (EDM) \cite{9925094} plays a critical role in this transformation using machine learning and data mining algorithms to extract valuable insights from educational data, with a focus on improving educational performance \cite{khan2021student, batool2023educational}. Student performance prediction represents a typical application of EDM, guiding teachers to develop instructional programs that are both effective and informed by data-driven insights \cite{xu2020student, 10065517, chen2023comparative}. 

Recent advances in machine learning (ML) have markedly advanced the prediction of student performance, exerting a profound and fundamental impact on subsequent components such as personalized learning, targeted interventions, and overall educational outcomes \cite{yu2024mapping, webb2021machine}. For example, Support Vector Machines (SVM), a machine learning method known for its high precision and robust ability to handle nonlinear models, has been pivotal in various scenarios related to student performance prediction, such as identifying at-risk and borderline students \cite{verma2022scalable}, as well as forecasting student grades \cite{chui2020predicting}. In addition, the work of Tomasevic et al. \cite{tomasevic2020overview} adequately compared classical advanced supervised ML techniques on the tasks of identifying high-risk dropouts and predicting final exam grades, confirming that using student engagement data and historical performance information could help the models achieve relatively high prediction accuracy in classification and regression tasks. Besides, Luan and Tsai \cite{luan2021review} reviewed various empirical studies on precision education, underscored the predominance of machine learning in predicting academic performance and dropout rates. This research illuminates the potential of these methods in aiding educators' decision-making, particularly through the analysis of learning patterns and prediction of students' needs. Additionally, Pallathadka et al. \cite{pallathadka2023classification} classified and predicted students' academic performance through the Naive Bayes \cite{peretz2024naive} and C4.5 \cite{chen2023design} algorithms, demonstrating the potential of educational data mining in improving the quality of education and reducing the dropout rate, and providing a reference for optimizing teaching. Moreover, Cheng et al. \cite{cheng2024evaluation} adopted multiple classifiers including Multilayer Perceptron (MLP) and eXtreme Gradient Boosting (XGBoost) \cite{ramraj2016experimenting}, and further combined optimization algorithms to improve the accuracy of predicting students' performance.

The aforementioned studies has confirmed the effectiveness of machine learning in predicting student performance. However, given the diversity of educational data and the variety of practical tasks, a comprehensive comparison of different machine learning methods is essential. Furthermore, existing studies have yet to fully harness the potential of prediction results. This underscores the necessity for improved integration of these methodologies within real teaching contexts, ensuring that the valuable insights derived from machine learning can be effectively leveraged to enhance educational strategies and ultimately improve educational outcomes \cite{pietsch2024leading, elen2024education}. 

To fully leverage the potential of advanced technologies such as machine learning in contemporary educational contexts, we propose a framework that integrates the outcomes of machine learning-based student performance prediction with tiered instruction to improve student outcomes in the target course. Firstly, original datasets are collected and subjected to detailed analysis and preprocessing. The data covers compulsory and specialized courses as specified in the syllabus, as well as some student behavioral attributes. Secondly, to enhance the accuracy of prediction, the model inputs are optimized through feature engineering in this study, and the performance of multiple machine learning algorithms in predicting students' grades is compared. Thirdly, the most effective method is employed for classification predictions in the target course and the results of these predictions are strategically utilized to formulate tiered teaching strategies. Finally, to validate the effectiveness of the proposed framework, a controlled experiment is designed. Specifically, students are categorized into different levels based on the classification results of the optimal model, and tiered instruction is implemented in the experimental class, while the control class maintains traditional teaching methods. Comparison with the control class and questionnaire analysis confirms the effectiveness of tiered instruction in improving student achievement in the experimental class. It is important to highlight that while this study is centered around a particular computer course, the proposed framework possesses a broader applicability, extending its relevance to various subjects and diverse educational levels.

This article is organized as follows. Related works is presented in section \uppercase\expandafter{\romannumeral 2}. The method of this paper is described in section \uppercase\expandafter{\romannumeral 3}. The results of the experiments and the discussion are presented in section \uppercase\expandafter{\romannumeral 4}. The concepts related to tiered instruction, specific implementation design and corresponding effects are presented in section \uppercase\expandafter{\romannumeral 5}. Finally, conclusion and future works are given in section \uppercase\expandafter{\romannumeral 6}.

\section{Related works}
\subsection{Machine learning in student performance prediction}
Research on predicting student performance using machine learning is a hot topic that continues to attract a wide range of researchers \cite{su2018exercise, silva2024identifying}. For instance, Bujang et al. \cite{bujang2021multiclass} conducted a comprehensive analysis of machine learning techniques and predicted the final grades of first semester based on the model achieving best accuracy. Similarly, Ko and Leu \cite{9144429} applied supervised and unsupervised machine learning techniques to identify the attributes of successful learners in a computer course, discovering that the Naïve Bayes classifier was the most effective for predicting student performance, and revealed the importance of weekly progress and self-efficacy beliefs in influencing final outcomes. Xu et al. \cite{9162494} gathered data describing student learning behavior and used it to test its predictability over 1/4, 1/2, and 3/4 semesters via multiple regression models, which experimentally proved to be more consistent and reliable as the course progressed. Said et al. \cite{ben2024early} proposed a ML-based method to predict undergraduate academic performance and dropout risk, focusing on early identification of students needing attention from the first semester due to academic weaknesses. It identifies key factors such as demographic, pre-admission, and academic, that influence academic performance, supporting educational decision-making. Accordingly, feature selection is an important research subject in machine learning and data mining, and the prediction performance of the model can be effectively improved by appropriate feature selection methods. 

To thoroughly investigate the impact of feature selection on the performance of different prediction models, Zhou et al. \cite{zhou2015performance} conducted a series of experiments, and the empirical results showed that the feature selection methods played a significant role in predicting student performance. Building on this, Peng et al. \cite{peng2023examining} employed Random forest to pinpoint the key information and communication technology related factors influencing reading performance in blended learning, highlighting AI's role in educational technology optimization. Talebi et al. \cite{talebi2024ensemble} performed feature selection by eliminating redundant features using Recursive Feature Elimination (RFE), and selecting the most predictive features based on feature importance scores from machine learning models, thereby optimizing the feature set. Despite these advancements, current research on predicting student performance with machine learning has largely failed to extended the application of these model-driven insights to real-world educational settings. As a result, bridging the gap between research findings and their practical implementation in educational contexts remains a pressing issue. 

\subsection{Tiered instruction}
The tiered instruction \cite{richards2007effects} refers to the design of multi-level learning content based on learners' existing skills and knowledge, aiming to support each student in engaging in effective learning at a level appropriate to their abilities, thereby optimizing their learning experience and achievement \cite{nichols2020teacher}. This innovative educational method is an innovative educational method commonly used in today's educational environment, has drawn significant attention. For instance, Pullen et al. \cite{pullen2010tiered} provided additional instruction to at-risk students through the design and use of tiered interventions in the general education classroom and assists students at risk for dyslexia. Additionally, Freeman-Green et al. \cite{freeman2018mathematics} effectively used the tiered approach at the students' secondary school level to enhance students' skills and abilities in a targeted manner according to the teaching requirements, and achieved excellent teaching results. Magableh et al. \cite{magableh2020effectiveness} investigated the effects of differentiated instruction on enhancing the overall academic achievement of Jordanian students. By employing tiered assignments and group-based teaching that adjusted instructional content according to students' abilities and interests, the study demonstrated a significant improvement in student performance. Vojinovic et al. \cite{vojinovic2020tiered} proposed a tiered lab programming model that helped students gradually overcome learning challenges through tiered assignments and deliberately set learning obstacles. The study demonstrated that this approach significantly enhanced student motivation and exam pass rates, with particularly notable results among previously unmotivated students. 

However, as one of the most effective classroom teaching methods, the tiered instruction has not been well integrated with modern technology to demonstrate better performance. In addition, although many existing machine learning methods have shown good performance in educational data mining, the predictive outcomes of the models have not been effectively utilized in actual teaching scenarios. Based on the existing researches, we combine the analysis results of machine learning with actual teaching scenarios by introducing tiered teaching method and propose a specific implementation plan, which provides a novel insight on the application of machine learning in modern education. To the best of our knowledge, this is the first work to combine machine learning methods with tiered instruction.

\section{Method}
\begin{figure*}[t]
    \centering
	\includegraphics[width=1\linewidth]{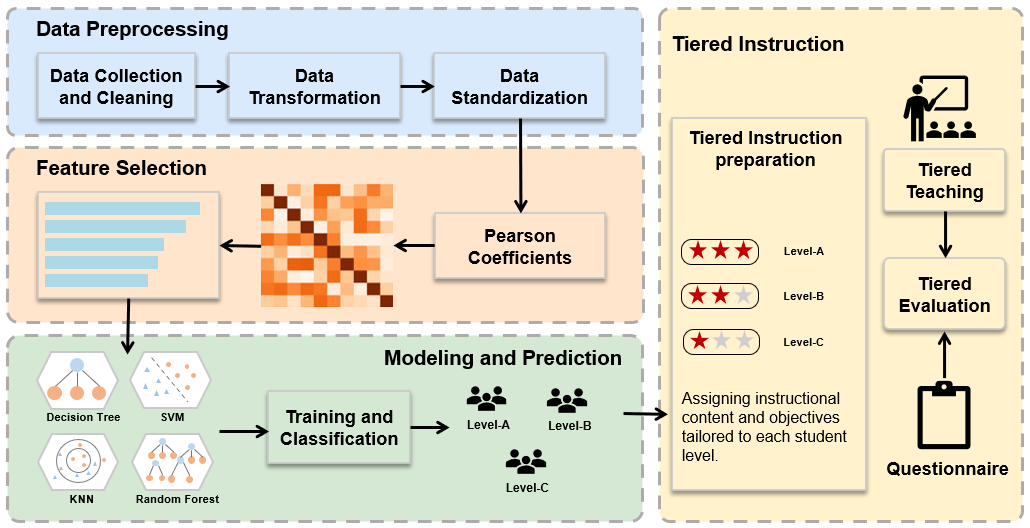}
    \caption{Research framework diagram of this work.}
    \label{fig_1}
\end{figure*}

The research diagram of this work is shown in Figure \ref{fig_1}, which consists of data gathering and preprocessing, feature correlation analysis and feature selection, modeling and prediction, and tiered instruction. First, we preprocess the original data, and perform feature selection by  correlation analysis. Furthermore, the selected features are are applied to five representative machine learning methods for prediction and the best-performing model is used to categorize the predicted results into three tiers, corresponding to different levels of students. This categorisation aims to integrate the tiered teaching methodology with actual teaching scenarios. Finally, the effectiveness of the teaching methodology is assessed through comparative analyses of performance with the control class and statistical evaluations of questionnaires.

\begin{table}[t]
\caption{Overview of the features and descriptions of the raw data gathered}
\centering
\begin{tabularx}{\linewidth}{p{2cm}p{3cm}X}
\hline
Attributes & Features & \makecell[ct]{Description} \\
\hline
Compulsory \newline Courses & Language \newline Mathematics \newline English \newline PE & Foundational courses required in the educational curriculum, essential for students' academic and health development \\
\hline
Specialized \newline Courses & Database \newline Java \newline \footnotesize{Computer Network} &  Specialized courses designed to provide in-depth knowledge in specific fields of study within the development program \\
\hline
Behavioral & Gender \newline Study Time \newline Attendance &  Attributes and behavioral patterns that can influence or reflect students' learning outcomes \\
\hline
Targeted & \small{The Principle of} \newline \small{Microcomputer} & Targeted courses that plan to implement new teaching methods\\
\hline
\end{tabularx}
\label{Table1}
\end{table}

\subsection{Data gathering and preprocessing}
The data gathered in this study originate from the computer program of a secondary vocational and technical school during the author's internship, which includes course grades of a total of 2,000 students from different grades over all the compulsory courses in the previous semester, such as language, mathematics, English, database, Java, computer network and PE. The above course achievement data combine with students' personal attributes such as gender, study time and attendance as input features, aiming to predict students' performance in the next semester in the principle of Microcomputer. 

Table \ref{Table1} summarizes the raw data gathered for this study, which is divided into four components according to feature attributes: compulsory courses, specialized courses, behavioral attributes, and target course. As required by the training program, language, mathematics, English and PE as compulsory courses, are essential for the establishment of the students' academic foundation and the promotion of their physical and mental health. Specialized courses cover databases, Java and computer networks and are designed to provide in-depth knowledge in specific areas of the program. Regarding behavioral attributes, data on gender, study time, and attendance are collected to assist in the analysis of how students' behavioral patterns affect learning outcomes. The Principle of Microcomputer is not only the target for predicting student achievement but also a benchmark for applying new teaching methods.

The original dataset contains missing values, outliers, and inconsistencies, and covers a variety of data types including course grades, students' personal characteristics, and behavioral attributes, which vary in value ranges and characteristics. Therefore, this study implements specific data preprocessing strategy for cleaning and transformation to improve data quality and ensure that the dataset is better adapted for subsequent classification algorithm applications.

To avoid unnecessary data wastage, missing data due to improper recording of school information are filled in with mean values. In the raw data, gender is presented in a non-numeric form, so males and females are numerically coded as 0 and 1, respectively, and further converted to a format suitable for model inputs by applying one-hot coding. Additionally, since the original data features take on different ranges of values, directly feeding them into the model might lead to an overemphasis on features with larger numerical values in the model analysis, reducing the impact of features with smaller values. Therefore, to preserve the balanced importance of each feature in the model, we have normalized the data, ensuring uniformity in the scale of all features. This approach enables a more thorough and accurate comprehensive analysis and assessment.

\subsection{Feature correlation analysis and feature selection}

\begin{figure}[t]
    \centering
	\includegraphics[width=0.6\linewidth]{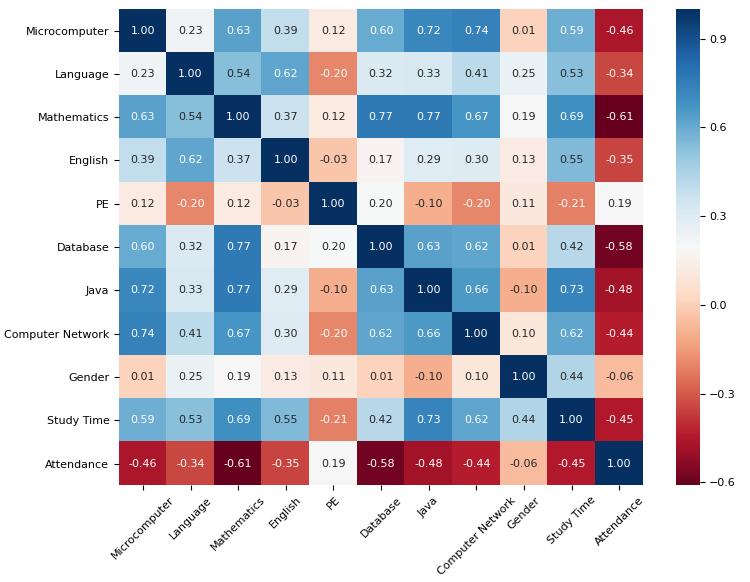}
    \caption{Heatmap of correlation between features.}
    \label{fig_2}
\end{figure}

For enhanced classification prediction, feature correlation analysis is a crucial step following data preprocessing. The correlation coefficient serves as a quantitative metric to assess the interrelationship between each pair of features, effectively describing the degree of correlation between them. Based on it, features with the strongest correlation to the target variable are selected, enabling the model to focus on the most influential variables, which significantly enhances the model's predictive reliability and accuracy. In this study, we employ the Pearson correlation coefficient \cite{cohen2009pearson} for analysis. A coefficient greater than 0 indicates a positive correlation, with values closer to 1 signifying stronger relationships. Conversely, values less than 0 denote negative correlations, with those nearing -1 indicating stronger inverse relationships. Values around 0 suggest negligible correlation. The heatmap visually represents the correlation degree between different features with different depths of colors, which is helpful for feature selection. The visualization is shown in Figure \ref{fig_2}.

The heatmap of Figure \ref{fig_2} shows the correlation between the performance scores in the target course 'Principle of Microcomputer' and various other features. In particular, Java and Computer Networks show a strong positive correlation of more than 70\% with it, indicating that an in-depth understanding of these technical subjects is critical to the target course. Correspondingly, Mathematics displays over a 60\% positive correlation, highlighting the importance of mathematical logic in comprehending the intricacies of microcomputer concepts. However, the lower correlation with courses such as language and English suggests that proficiency in professional skills and knowledge outweighs linguistic abilities in contributing to academic success in technical fields. Additionally, the positive correlation (59\%) between study time (hours) and Principle of Microcomputer scores suggests that time invested in after-class study positively influences academic performance. Conversely, the negative correlation with attendance (times of being late) implies that students with higher absenteeism may lack motivation and good study habits, which adversely affect their grades. In summary, these analyses not only deepen the understanding of the learning outcomes in the target course but also provide empirical support for educators to formulate more effective teaching strategies, particularly in areas of after-class study support and attendance management. To prevent overfitting and reduce computational load, this study has selected Mathematics, Database, Java, and Computer Network course scores, along with study time and attendance rate, as the input variables for the predictive models.

In addition, the correlations of other features presented in the heatmap are also insightful. The correlations among the three fundamental courses, Language, Mathematics, and English, suggest that students’ performance in these subjects tends to influence one another. Additionally, the positive correlation between these foundational courses and specialized subjects such as Database, Java, and Computer Network underscores the role of foundational education in supporting the mastery of professional knowledge. Furthermore, the high correlation among specialized courses indicates an interdependence of different professional knowledge points within the field, sharing core concepts and skills. However, the correlation between PE performance, gender, and other characteristics is low, reflecting the limited direct impact of non-academic factors on learning outcomes. Moreover, the negative correlation between study time and the number of late arrivals may suggest that students who invest more time in after-class studies generally exhibit a more serious attitude towards learning. This finding further emphasizes the link between students' study habits and their classroom performance. Ultimately, these detailed descriptions not only reveal the interrelationships between different subjects but also provide insights into students’ learning behaviors and attitudes, offering data support for the formulation of educational interventions and personalized teaching strategies.

\subsection{Modeling and prediction}
In this subsection, we implement machine learning algorithms using the scikit-learn library in Python to apply five classical classification models for performance prediction, including K-Nearest Neighbors (KNN) \cite{amra2017students}, Naive Bayes, SVM, Decision Tree, and Random Forest. The ultimate goal is to classify students into three categories. While deep learning methods may offer greater accuracy, this study focuses on predictions and adjustments in teaching strategies based on a small dataset collected over a short period. Therefore, in such scenarios, classic machine learning methods are more appropriate.

KNN is one of the most commonly used classification algorithms in machine learning methods, and it performs classification prediction by measuring the distance between different features. When performing classification by KNN, the category of a new samples can be determined based on the nearest K samples to it. The model is fast in training and insensitive to outliers, making it an important candidate when making classification. Many researchers have applied KNN algorithm to make predictions \cite{maghari2018prediction}\cite{yekun2021student}. In particular, Maghari used modified KNN classifiers, including cosine KNN, cubic KNN, and weighted KNN, to predict students' grades and achieved promising performance \cite{maghari2018prediction}. In our experiments, a grid search is employed to identify the optimal value of K.

Naive Bayes is a simple probabilistic technique based on Bayes' theorem which uses the assumption of conditional independence of features, i.e., each feature is assumed to independently influence the classification result \cite{mengash2020using}. It performs well on small-scale data and can handle multi-classification tasks, and it has been proven to achieve excellent performance in various studies \cite{priya2020novel}\cite{albahli2024efficient}.

The basic principle of SVM is to solve for the separated hyperplane that correctly partitions the data set and maximizes the geometric separation. SVM is well suited for the study of small data sets, and is often used in the study of some classification problems due to its advantages such as strong interpretability and effectiveness \cite{liu2021projection}. Also, the final decision function of SVM is determined by only a small number of support vectors, which in a sense avoids curse of dimensionality.

Decision tree is a simple and widely used classifier that can efficiently classify unknown data by building tree model from training data. It uses a recursive approach to construct a tree with the fastest decreasing entropy value. Generally, three basic steps are included: feature selection, decision tree generation, and decision tree pruning. Decision tree is widely used by researchers in classification studies for its interpretability, descriptiveness and efficiency \cite{silva2024identifying}.

Random forest is an ensemble learning based method, which is the construction and aggregation of multiple base learners to accomplish a machine learning task. Specifically, random forest uses multiple decision trees to train and make predictions, and merge them together to obtain more accurate and stable results. Compared with other algorithms, random forest has excellent accuracy and can handle input samples with high-dimensional features. Furthermore, it can evaluate the importance of each feature in the classification problem, so it has a wide range of application prospects. 

\section{Experimental results}
In this section, we assess the performance of the five predictive models utilizing four key evaluation metrics: Accuracy, Recall, Precision, and F-measure. We then analyze and discuss the outcomes derived from these experimental evaluations.

\subsection{Evaluation metrics}
In this work, we regard the grade prediction for the target course as a classification task, and the four metrics are selected to evaluate the predictive performance of the models.

Accuracy is the percentage of correctly predicted results and is measured by (\ref{eq1})
\begin{equation}
    Accuracy = (TP+TN)/(TP+TN+FP+FN)
    \label{eq1}
\end{equation}

Recall is the percentage of positives that are correctly predicted as positive and is measured by (\ref{eq2})
\begin{equation}
    Recall = TP/(TP+FN)
    \label{eq2}
\end{equation}

Precision is the percentage of correct positive observations and is measured by (\ref{eq3})
\begin{equation}
    Precision = TP/(TP+FP)
    \label{eq3}
\end{equation}

F-measure is a weighted summed average of recall and precision and is measured by (\ref{eq4})
\begin{equation}
    F-measure = 2\times Recall\times Precision/(Recall+Precision)
    \label{eq4}
\end{equation}
where $TP$ denotes True Positive, $TN$ denotes True Negative, $FP$ denotes False Positive, and $FN$ denotes False Negative.

\subsection{Analysis and discussion for experimental results}
The five classification techniques mentioned above are evaluated using the selected dataset split where 60\% is allocated for model training, and 20\% each is used for validation and testing, respectively. The models are trained on the training set using the 10-fold cross-validation method. Parameter optimization is conducted by generating learning curves and utilizing grid search to identify the parameters that yield the best outcomes on the validation set. Ultimately, classification predictions are performed on the test set. The performance comparison is presented in Table \ref{Table1}, which includes calculated values for Accuracy, Precision, Recall, and F-measure.

\begin{figure}[t]
    \centering
	\includegraphics[width=.8\linewidth]{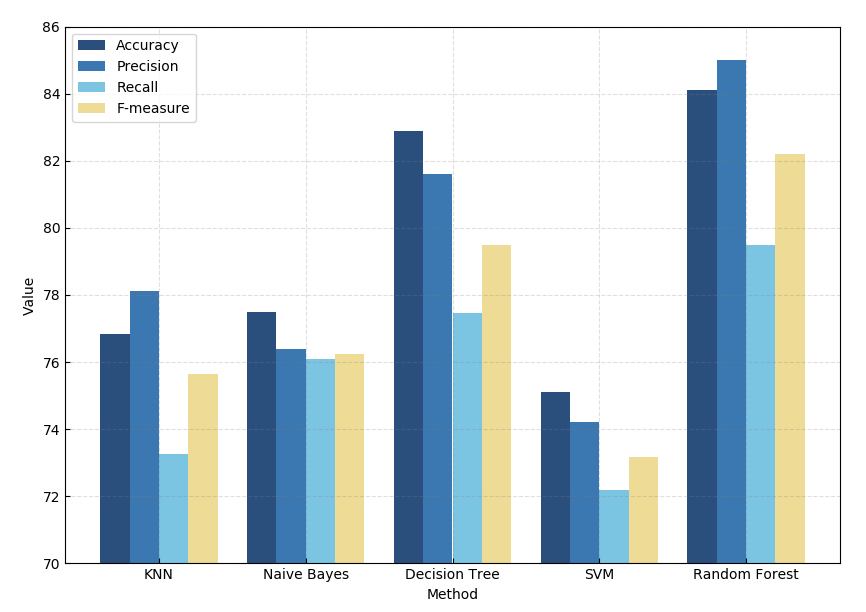}
    \caption{Visualization of classification performance of five classifiers.}
    \label{fig_3}
\end{figure}
To visualize the performance effects of different models, Figure \ref{fig_3} shows the visualization of the classification performance of five classifiers. It can be more intuitively seen that random forest technique outperforms other techniques in four performance metrics, where the accuracy rate can reach 84.10\%, the precision rate reaches 85.01\%, the recall rate is 79.49\% and the F-measure rate is 82.20\%. Decision tree with the accuracy rate of 82.90\%, is second only to random forest. The overall performance of KNN and Naive Bayes is very similar, while the results of Naive Bayes are more stable. In addition, SVM achieves the weakest performance. 

\begin{table}[t]
\caption{Performance comparison for different methods}
\centering
\begin{tabular}{ccccc}
\hline
Method & Accuracy & Precision & Recall & F-measure\\
\hline
KNN & 76.85 & 78.12 & 73.25 & 75.64\\
Naive Bayes & 77.50 & 76.40 & 76.10 & 76.25\\
Decision Tree & 82.90 & 81.60 & 77.45 & 79.48\\
SVM & 75.10 & 74.20 & 72.20 & 73.18\\
Random Forest & \textbf{84.10} & \textbf{85.01} & \textbf{79.49} & \textbf{82.20}\\
\hline
\end{tabular}
\label{Table2}
\end{table}

In fact, the classical SVM model is suitable for application to binary classification problems, with the core idea being to find a hyperplane that maximally distinguishes between two classes of data points, while the performance degrades when treating multi-classification problems. The KNN relies heavily on training data and greatly affects the KNN classification performance when the distribution of sample categories is unbalanced. The Naive Bayes model predicts student performance by assuming that the features are independent, while there is a degree of correlation between the features in the actual data, which causes the poor performance of the model. Decision tree divides the data based on information gain to construct a tree model for providing easily interpretable prediction results and achieving better classification performance. Random Forest is the integration of a series of Decision Trees, so it can further improve the classification performance based on the Decision Trees, and the voting mechanism can eliminate the overfitting. Thus, Random Forest exhibits the most effective performance in predicting student performance.

The experiments and analysis presented confirm that machine learning is capable of predicting student grades effectively. However, the application of these predictions in educational settings has been limited. As noted in the Section II, our study addresses this gap by linking machine learning outcomes with actual teaching practices through tiered instruction. This approach yields a pragmatic plan for applying machine learning insights to enhance educational strategies.

\section{Tiered instruction}
Most research only reaches the point of obtaining classification results for student grades and does not extend further into practical teaching applications, thus lacking in practical educational value. Traditional teaching methods struggle to accommodate students of varying abilities, often failing to achieve optimal teaching outcomes. Tiered teaching strategies are being explored as innovative approaches to foster more effective and inclusive learning environments. Thus, we employ the best-performing Random Forest algorithm to predict the performance classification results of 50 students from each of Class 1 (tiered class) and Class 2 (traditional class) in the principle of Microcomputer course. Utilizing these classification results, we effectively divide the students into three distinct levels. This stratification serves as a foundational step towards the implementation of tiered teaching methods. By applying tiered instruction in one class and traditional teaching in another to form a control experiment, we validate the actual effectiveness of the tiered teaching method by comparing the differences in the end-of-term exam grade distributions between the two classes.

Based on the effective classification of students by machine learning techniques and the specific requirements of the training program, this study has developed customized teaching strategies for different learning levels, as shown in Table \ref{Table3}. The teaching objectives for level C students aim to establish a solid knowledge foundation with content presented through intuitive multimedia materials to avoid excessive theorization. The corresponding assignment design emphasizes the practice of basic skills, along with understanding and applying classroom-taught content. Level B students need to expand their ability to solve practical problems through case studies and inquiry learning content and related assignments, and adapt to the requirements of the profession on students' practical skills. For level A students, the teaching objectives emphasize the development of innovative abilities, therefore it is essential to design teaching content integrating cutting-edge technologies and strategically incorporates open-ended questions and assignments for independent learning. This approach ensures that students not only master theoretical aspects but also excel in applying their acquired knowledge to practical scenarios.

\begin{table*}
\caption{Tiered instructional design strategies based on machine learning student classification}
\centering
\begin{tabularx}{\linewidth}{cp{3.2cm}X}
\hline
 \footnotesize{Level} & \footnotesize{Instructional Design} & \footnotesize\makecell[ct]{Specific requirements for instructional design}\\
\hline
 \multirow{3}{*}{C} & \footnotesize{Teaching objective} & \footnotesize{Adhere to the principle of consolidating fundamentals, focusing on the mastery of basic knowledge and core skills.}\\
 
 & \footnotesize{Teaching Content} & \footnotesize{Present conceptual content using a variety of media, avoiding complex and abstract theories.}\\
 
 & \footnotesize{Assignment} & \footnotesize{Focus on the practice of basic skills, provide fundamental exercises to reinforce classroom content.}\\
\hline
\multirow{3}{*}{B} & \footnotesize{Teaching objective} & \footnotesize{Follow the principle of integrated application to cultivate the ability to apply knowledge and problem-solving skills.}\\

 & \footnotesize{Teaching Content} & \footnotesize{In conjunction with case studies, introducing investigative content on top of fundamental knowledge.}\\
 
 & \footnotesize{Assignment} & \footnotesize{Design projects that combine theory and practice, applying acquired knowledge to solve real-world problems.}\\
 \hline
 \multirow{3}{*}{A} & \footnotesize{Teaching objective} & \footnotesize{Guided by the principles of creativity, develop independent thinking skills, a spirit of inquiry, and self-driven abilities.}\\
 
 & \footnotesize{Teaching Content} & \footnotesize{Showcases the latest technology trends and emphasizes the integration of interdisciplinary knowledge.}\\
 
 & \footnotesize{Assignment} & \footnotesize{Design open-ended questions and promote independent research and self-directed learning.}\\
\hline
\end{tabularx}
\label{Table3}
\end{table*}

In the practice of tiered instruction in the target course Principle of Microcomputer, the 50 students in the experimental class are correspondingly classified into three levels, so as to achieve student stratification. Specifically, Level A refers to students who are likely to score 80 or above in the target course, Level B refers to students scoring between 60-79, and Level C refers to students scoring below 60. It is important to emphasize that student stratification is not meant to judge students, but to provide targeted guidance for students at different levels, aiming to improve their grades while protecting their self-esteem.

In accordance with the tiered teaching strategy outlined in Table \ref{Table3} and the requirements of the teaching syllabus, this study has developed specific educational objectives, content, and assignments for students at different levels in the experimental class. Adhering to a principle of progressing from simpler to more complex concepts, the design is structured as follows.

\noindent\textbf{Level C.} For Level C students, the objective is to master the basic components and operating principles of microcomputers. Consequently, the teaching content includes the hardware composition of personal computers, the functions of the Basic Input Output System (BIOS), and fundamental concepts of operating systems. Corresponding assignments are designed to enhance recognition of hardware component functions. 

\noindent\textbf{Level B.} For Level B students, who are expected to gain a comprehensive understanding and application of microcomputer systems, the course content extends to diagnosing and handling common hardware malfunctions and basic network setup. Assignments for these students involve solving specific hardware malfunction cases. 

\noindent\textbf{Level A.} The teaching goal for Level A students is to improve practical skills and innovative thinking. Therefore, the content expands to include programming techniques and project practice skills, with assignments that require the development of simple applications and innovative solution proposals.

The tiered teaching strategy implemented in this study aligns with the needs of modern education, offering personalized education to students with varying needs and levels of ability. Initially, the classification results from the Random Forest effectively stratify students, enabling teachers to tailor educational objectives, content, and assignments for each learning level. This approach enhances the specificity and effectiveness of teaching, thereby respecting the differences among students and overcoming the limitations of traditional teaching methods. Overall, this tiered teaching strategy provides an effective means to meet the specific needs of different students, creating favorable conditions for improving teaching quality and learning outcomes.

\subsection{Effectiveness of Teaching Practices}
A semester-long tiered teaching method was employed to teach the Principles of Microcomputer in Class 1 of the vocational secondary school based on the classification prediction results of machine learning. Concurrently, the traditional teaching mode was used in the 'Principles of Microcomputer' course of the Class 2 of computer application majors, thus forming a controlled experiment. To objectively evaluate the practical effects of the two teaching modes, a test paper was designed at the end of the semester that comprehensively covered all the key contents of the course for the semester. The data from the exam results of both classes were collected, organized, and analyzed to visually compare the two teaching methods. The recovery rate of valid test papers in each class is 100\%, and the efficiency rate is 100\%. Table \ref{Table4} and Table \ref{Table5} respectively present the comparison of machine learning prediction results and actual teaching effectiveness in Class 1 and Class 2.

\begin{table}[t]
\caption{Comparison of Machine Learning Prediction Results and Actual Teaching Effectiveness in Class 1}
\centering
\begin{tabular}{cccc}
\hline
 & Level A (80-100) & Level B (60-79) & Level C ($<$60)\\
\hline
Predicted & 16(32\%) & 24(48\%) & 10(20\%)\\
Actual & 22(44\%) & 26(52\%) & 2(4\%)\\
\hline
\end{tabular}
\label{Table4}
\end{table}

\begin{table}[t]
\caption{Comparison of Machine Learning Prediction Results and Actual Teaching Effectiveness in Class 2}
\centering
\begin{tabular}{cccc}
\hline
 & Level A (80-100) & Level B (60-79) & Level C ($<$60)\\
\hline
Predicted & 15(30\%) & 26(52\%) & 9(18\%)\\
Actual & 17(34\%) & 25(50\%) & 8(16\%)\\
\hline
\end{tabular}
\label{Table5}
\end{table}

\begin{figure}[t]
    \centering
    \begin{subfigure}{0.45\linewidth}
    		\centering    		\includegraphics[width=1\linewidth]{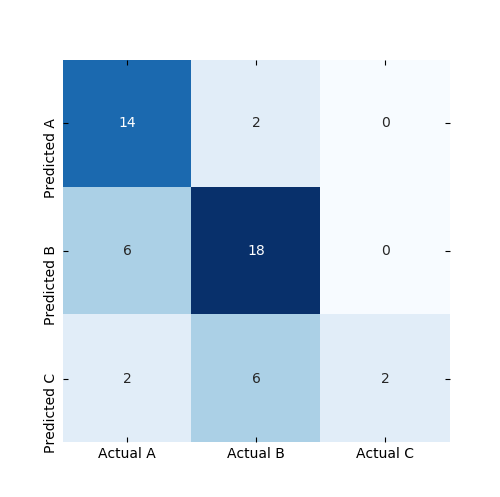}
    		\caption{}
    		\label{fig4_a}
    \end{subfigure}   
  \begin{subfigure}{0.45\linewidth}
		\centering
		\includegraphics[width=1\linewidth]{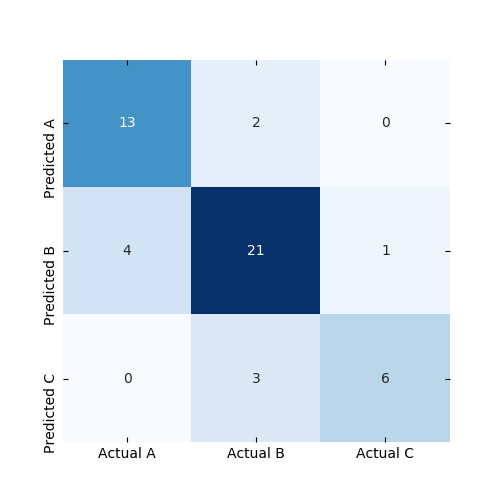}
		\caption{}
		\label{fig4_b}
	\end{subfigure}
    \caption{Confusion matrixes of the predicted and actual student counts for different levels in Class 1 (a) and Class 2 (b).}
    \label{fig_4}
\end{figure}

The small differences between the predictions of the Random Forest model and the actual teaching results shown in Table \ref{Table5} validate the accuracy of the model in predicting grades in the Principles of Microcomputer course. Since the Random Forest model is trained using grades obtained by students under the traditional teaching mode, and the actual teaching outcomes are also derived from this mode, the classification results presented in Table \ref{Table5} further demonstrate the accuracy of our predictions.

Based on the above analyses, Table \ref{Table4} shows the superiority of the tiered teaching method compared to the traditional teaching model. The classification prediction results of Random Forest show that only 16 students in the class taught under the traditional teaching mode can get A level at the end of the semester, and the number of failing students is 10. However, after the actual implementation of the tiered teaching mode, the number of students obtaining an A grade in the class reached 22, accounting for 44\% of the total class size, and the actual number of failing students in the class decreased from the predicted 10 to 2. In addition, there has been an increase in the number of students at the B level. Thus, this further indicates that students at this level have a better foundation and should be given more attention and support to motivate them to pursue higher learning goals.

Moreover, according to Figure \ref{fig_4} (a), 14 of the students who actually receive A grades are correctly predicted, making up the majority of the 16 anticipated to achieve this grade. Another 6 students exceed expectations by moving up from predicted B grades, and 2 from predicted C grades. Additionally, 26 students actually achieve B grades, with 18 matching their predictions and 6 advancing from C grades. All students receiving C grades align with the predictions. In contrast, Figure \ref{fig_4} (b) shows minimal discrepancies between predicted and actual grade distributions, with no significant shifts between levels. This highlights the accuracy of the model and the effectiveness of the tiered instruction in raising student achievement, especially for those students who are initially underestimated.

\subsection{Analysis of questionnaires}
When adopting a new teaching method, it is crucial to obtain students' feedback, which helps teachers to grasp students' actual feelings about the tiered teaching method in specialized courses, and make necessary adjustments and optimization of teaching strategies accordingly. After a semester of application, at the end of the term, we distribute 50 relevant questionnaires to Class 1 students and collect 50 valid questionnaires, achieving a 100\% recovery rate and validity. From the statistical analysis of the questionnaire data, we can make the following summary of several key issues:

\begin{figure}[t]
    \centering	
    \includegraphics[width=0.6\linewidth]{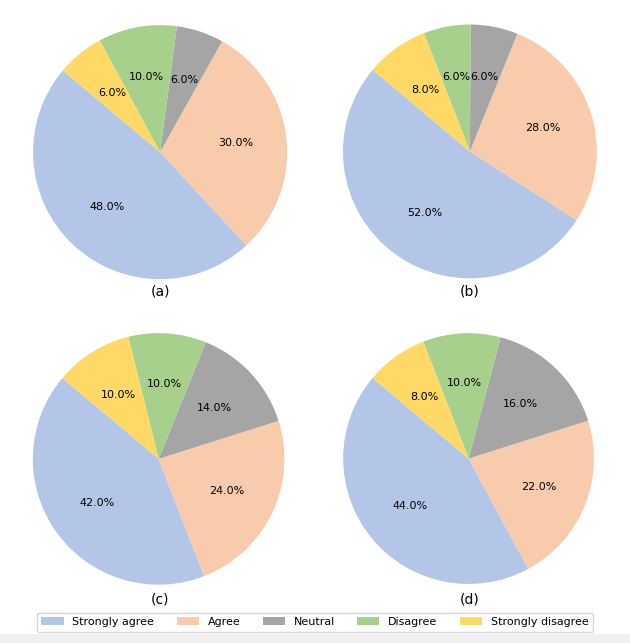}
    \caption{Four statistical cases of questions from the survey questionnaire.}
    \label{fig_5}
\end{figure}

The pie charts Figure \ref{fig_5} present a statistical overview of students' attitudes toward the implementation of tiered teaching methods in the Principle of Microcomputer course, focusing on four representative questions: "The implementation of tiered teaching methods has increased my interest in learning," "My learning objectives are clearer," "The teacher's student tiering is reasonable," and "The teacher's use of teaching resources is appropriate." The data indicate a positive student response to the tiered teaching methods in enhancing learning interest and clarifying learning objectives, reflecting the methods' effectiveness in boosting students' intrinsic motivation and learning efficiency. Additionally, the majority of students acknowledge the fairness of the teachers' tiering strategy, suggesting that tiered teaching is perceived as equitably recognizing and meeting the diverse abilities and needs of students. The favorable evaluations of the use of teaching resources further underscore the pivotal role of teachers in the application of tiered teaching, implying that proper resource alignment can enhance teaching outcomes. However, the presence of a minority of students who chose neutral, disagree, or strongly disagree responses indicates room for improvement in the application of tiered teaching methods. This feedback may point to the need for more refined tiered instructional strategies to increase support for specific groups of students.
 
The prevailing educational paradigm emphasizes a student-centered approach, rendering student feedback on tiered instruction invaluable for evaluating pedagogical efficacy, which is instrumental in demonstrating the effectiveness of tiered teaching in meeting diverse learning requirements and fostering intrinsic motivation. Nonetheless, the existence of dissenting opinions, albeit from a minority, highlights the necessity for continual refinement of teaching strategies. The insights gleaned from student feedback thus serve as a compass for educational adjustments, paving the way for more personalized, effective, and student-centered learning environments.

\section{Conclusion and future works}

This research aims to equip modern educators with innovative teaching strategies by integrating machine learning techniques into tiered instruction framework. Specifically, we preprocess the collected data to enhance the accuracy and efficiency of subsequent predictive analyses. Given that original data often exhibit information redundancy due to diverse features, we employ correlation analysis for feature selection to obtain more related features with the target variable. Moreover, we explore and discuss five representative machine learning methods, among which Random Forest demonstrates notably promising performance. Building upon these results, we apply the tiered instruction approach in alignment with the predicted classification outcomes. This enables us to tailor teaching objectives and content to meet the specific needs of students at different learning levels, thereby fostering a more effective and personalized educational environment. Finally, the comparison of teaching outcomes with the control class validates the efficacy of the tiered teaching method, and the statistical results of the survey questionnaire indicate a high level of student satisfaction. 

It is noteworthy that although the context of this study is within a specific computer course, the method proposed here can be applied to other subjects or different grade levels. In the future, we aim to incorporate more advanced techniques, such as deep learning, to enhance the effectiveness of student performance prediction. Finally, exploring how to integrate modern technology into practical teaching scenarios remains a question of significant interest worthy of further study. 

\bibliographystyle{unsrt}
\bibliography{ref}

\begin{thebibliography}{10}

\bibitem{9925094}
Galina Deeva, Johannes De~Smedt, and Jochen De~Weerdt.
\newblock Educational sequence mining for dropout prediction in moocs: Model building, evaluation, and benchmarking.
\newblock {\em IEEE Transactions on Learning Technologies}, 15(6):720--735, 2022.

\bibitem{khan2021student}
Anupam Khan and Soumya~K Ghosh.
\newblock Student performance analysis and prediction in classroom learning: A review of educational data mining studies.
\newblock {\em Education and information technologies}, 26:205--240, 2021.

\bibitem{batool2023educational}
Saba Batool, Junaid Rashid, Muhammad~Wasif Nisar, Jungeun Kim, Hyuk-Yoon Kwon, and Amir Hussain.
\newblock Educational data mining to predict students' academic performance: A survey study.
\newblock {\em Education and Information Technologies}, 28(1):905--971, 2023.

\bibitem{xu2020student}
Zhuojia Xu, Hua Yuan, and Qishan Liu.
\newblock Student performance prediction based on blended learning.
\newblock {\em IEEE Transactions on Education}, 64(1):66--73, 2020.

\bibitem{10065517}
Fei Wang, Zhenya Huang, Qi~Liu, Enhong Chen, Yu~Yin, Jianhui Ma, and Shijin Wang.
\newblock Dynamic cognitive diagnosis: An educational priors-enhanced deep knowledge tracing perspective.
\newblock {\em IEEE Transactions on Learning Technologies}, 16(3):306--323, 2023.

\bibitem{chen2023comparative}
Yawen Chen and Linbo Zhai.
\newblock A comparative study on student performance prediction using machine learning.
\newblock {\em Education and Information Technologies}, 28(9):12039--12057, 2023.

\bibitem{yu2024mapping}
Ji~Hyun Yu, Devraj Chauhan, Rubaiyat~Asif Iqbal, and Eugene Yeoh.
\newblock Mapping academic perspectives on ai in education: trends, challenges, and sentiments in educational research (2018--2024).
\newblock {\em Educational technology research and development}, pages 1--29, 2024.

\bibitem{webb2021machine}
Mary~E Webb, Andrew Fluck, Johannes Magenheim, Joyce Malyn-Smith, Juliet Waters, Michelle Desch{\^e}nes, and Jason Zagami.
\newblock Machine learning for human learners: opportunities, issues, tensions and threats.
\newblock {\em Educational Technology Research and Development}, 69(4):2109--2130, 2021.

\bibitem{verma2022scalable}
Swati Verma, Rakesh~Kumar Yadav, and Kuldeep Kholiya.
\newblock A scalable machine learning-based ensemble approach to enhance the prediction accuracy for identifying students at-risk.
\newblock {\em International Journal of Advanced Computer Science and Applications}, 13(8), 2022.

\bibitem{chui2020predicting}
Kwok~Tai Chui, Dennis Chun~Lok Fung, Miltiadis~D Lytras, and Tin~Miu Lam.
\newblock Predicting at-risk university students in a virtual learning environment via a machine learning algorithm.
\newblock {\em Computers in Human behavior}, 107:105584, 2020.

\bibitem{tomasevic2020overview}
Nikola Tomasevic, Nikola Gvozdenovic, and Sanja Vranes.
\newblock An overview and comparison of supervised data mining techniques for student exam performance prediction.
\newblock {\em Computers \& education}, 143:103676, 2020.

\bibitem{luan2021review}
Hui Luan and Chin-Chung Tsai.
\newblock A review of using machine learning approaches for precision education.
\newblock {\em Educational Technology \& Society}, 24(1):250--266, 2021.

\bibitem{pallathadka2023classification}
Harikumar Pallathadka, Alex Wenda, Edwin Ramirez-As{\'\i}s, Maximiliano As{\'\i}s-L{\'o}pez, Judith Flores-Albornoz, and Khongdet Phasinam.
\newblock Classification and prediction of student performance data using various machine learning algorithms.
\newblock {\em Materials today: proceedings}, 80:3782--3785, 2023.

\bibitem{peretz2024naive}
Or~Peretz, Michal Koren, and Oded Koren.
\newblock Naive bayes classifier--an ensemble procedure for recall and precision enrichment.
\newblock {\em Engineering Applications of Artificial Intelligence}, 136:108972, 2024.

\bibitem{chen2023design}
Xinxin Chen.
\newblock Design and research of mooc teaching system based on tg-c4. 5 algorithm.
\newblock {\em Systems and Soft Computing}, 5:200064, 2023.

\bibitem{cheng2024evaluation}
Biqian Cheng, Yuping Liu, and Yunjian Jia.
\newblock Evaluation of students' performance during the academic period using the xg-boost classifier-enhanced aeo hybrid model.
\newblock {\em Expert Systems with Applications}, 238:122136, 2024.

\bibitem{ramraj2016experimenting}
Santhanam Ramraj, Nishant Uzir, R~Sunil, and Shatadeep Banerjee.
\newblock Experimenting xgboost algorithm for prediction and classification of different datasets.
\newblock {\em International Journal of Control Theory and Applications}, 9(40):651--662, 2016.

\bibitem{pietsch2024leading}
Marcus Pietsch and Dana-Kristin Mah.
\newblock Leading the ai transformation in schools: it starts with a digital mindset.
\newblock {\em Educational technology research and development}, pages 1--27, 2024.

\bibitem{elen2024education}
Jan Elen and Fien Depaepe.
\newblock Education and technology: elements of a relevant, comprehensive, and cumulative research agenda.
\newblock {\em Educational technology research and development}, pages 1--14, 2024.

\bibitem{su2018exercise}
Yu~Su, Qingwen Liu, Qi~Liu, Zhenya Huang, Yu~Yin, Enhong Chen, Chris Ding, Si~Wei, and Guoping Hu.
\newblock Exercise-enhanced sequential modeling for student performance prediction.
\newblock In {\em Proceedings of the AAAI Conference on Artificial Intelligence}, volume~32, 2018.

\bibitem{silva2024identifying}
MPRIR Silva, RAHM Rupasingha, and BTGS Kumara.
\newblock Identifying complex causal patterns in students’ performance using machine learning.
\newblock {\em Technology, Pedagogy and Education}, 33(1):103--119, 2024.

\bibitem{bujang2021multiclass}
Siti Dianah~Abdul Bujang, Ali Selamat, Roliana Ibrahim, Ondrej Krejcar, Enrique Herrera-Viedma, Hamido Fujita, and Nor Azura~Md Ghani.
\newblock Multiclass prediction model for student grade prediction using machine learning.
\newblock {\em IEEE Access}, 9:95608--95621, 2021.

\bibitem{9144429}
Chia-Yin Ko and Fang-Yie Leu.
\newblock Examining successful attributes for undergraduate students by applying machine learning techniques.
\newblock {\em IEEE Transactions on Education}, 64(1):50--57, 2021.

\bibitem{9162494}
Zhuojia Xu, Hua Yuan, and Qishan Liu.
\newblock Student performance prediction based on blended learning.
\newblock {\em IEEE Transactions on Education}, 64(1):66--73, 2021.

\bibitem{ben2024early}
Mouna Ben~Said, Yessine Hadj~Kacem, Abdulmohsen Algarni, and Atef Masmoudi.
\newblock Early prediction of student academic performance based on machine learning algorithms: A case study of bachelor’s degree students in ksa.
\newblock {\em Education and Information Technologies}, 29(11):13247--13270, 2024.

\bibitem{zhou2015performance}
Ligang Zhou, Dong Lu, and Hamido Fujita.
\newblock The performance of corporate financial distress prediction models with features selection guided by domain knowledge and data mining approaches.
\newblock {\em Knowledge-Based Systems}, 85:52--61, 2015.

\bibitem{peng2023examining}
Yi~Peng, Yanyu Wang, and Jie Hu.
\newblock Examining ict attitudes, use and support in blended learning settings for students’ reading performance: Approaches of artificial intelligence and multilevel model.
\newblock {\em Computers \& Education}, 203:104846, 2023.

\bibitem{talebi2024ensemble}
Kowsar Talebi, Zeinab Torabi, and Negin Daneshpour.
\newblock Ensemble models based on cnn and lstm for dropout prediction in mooc.
\newblock {\em Expert Systems with Applications}, 235:121187, 2024.

\bibitem{richards2007effects}
MRE Richards and Stuart~N Omdal.
\newblock Effects of tiered instruction on academic performance in a secondary science course.
\newblock {\em Journal of advanced academics}, 18(3):424--453, 2007.

\bibitem{nichols2020teacher}
Janet~Alys Nichols, William~D Nichols, and William~H Rupley.
\newblock Teacher efficacy and attributes on the implementation of tiered instructional frameworks.
\newblock {\em International Journal of Evaluation and Research in Education}, 9(3):731--742, 2020.

\bibitem{pullen2010tiered}
Paige~C Pullen, Elizabeth~D Tuckwiller, Timothy~R Konold, Katrina~L Maynard, and Michael~D Coyne.
\newblock A tiered intervention model for early vocabulary instruction: The effects of tiered instruction for young students at risk for reading disability.
\newblock {\em Learning Disabilities Research \& Practice}, 25(3):110--123, 2010.

\bibitem{freeman2018mathematics}
Shaqwana Freeman-Green, Julie Person, and Chris O'Brien.
\newblock Mathematics instruction for secondary students with learning disabilities in the era of tiered instruction.
\newblock {\em Insights into Learning Disabilities}, 15(2):175--194, 2018.

\bibitem{magableh2020effectiveness}
Ibrahim Suleiman~Ibrahim Magableh and Amelia Abdullah.
\newblock On the effectiveness of differentiated instruction in the enhancement of jordanian students' overall achievement.
\newblock {\em International Journal of Instruction}, 13(2):533--548, 2020.

\bibitem{vojinovic2020tiered}
Oliver Vojinovic, Vladimir Simic, Ivan Milentijevic, and Vladimir Ciric.
\newblock Tiered assignments in lab programming sessions: Exploring objective effects on students’ motivation and performance.
\newblock {\em IEEE Transactions on Education}, 63(3):164--172, 2020.

\bibitem{cohen2009pearson}
Israel Cohen, Yiteng Huang, Jingdong Chen, Jacob Benesty, Jacob Benesty, Jingdong Chen, Yiteng Huang, and Israel Cohen.
\newblock Pearson correlation coefficient.
\newblock {\em Noise reduction in speech processing}, pages 1--4, 2009.

\bibitem{amra2017students}
Ihsan A~Abu Amra and Ashraf~YA Maghari.
\newblock Students performance prediction using knn and na{\"\i}ve bayesian.
\newblock In {\em 2017 8th international conference on information technology (ICIT)}, pages 909--913. IEEE, 2017.

\bibitem{maghari2018prediction}
Ashraf Maghari.
\newblock Prediction of student's performance using modified knn classifiers.
\newblock In {\em Alfere, SS, \& Maghari, AY (2018). Prediction of Student's Performance Using Modified KNN Classifiers. In The First International Conference on Engineering and Future Technology (ICEFT 2018)}, pages 143--150, 2018.

\bibitem{yekun2021student}
Ephrem~Admasu Yekun and Abrahaley~Teklay Haile.
\newblock Student performance prediction with optimum multilabel ensemble model.
\newblock {\em Journal of Intelligent Systems}, 30(1):511--523, 2021.

\bibitem{mengash2020using}
Hanan~Abdullah Mengash.
\newblock Using data mining techniques to predict student performance to support decision making in university admission systems.
\newblock {\em Ieee Access}, 8:55462--55470, 2020.

\bibitem{priya2020novel}
K~Lakshmi Priya, Mourya Sai Charan~Reddy Kypa, Muchumarri Madhu~Sudhan Reddy, and G~Ram~Mohan Reddy.
\newblock A novel approach to predict diabetes by using naive bayes classifier.
\newblock In {\em 2020 4th International Conference on Trends in Electronics and Informatics (ICOEI)(48184)}, pages 603--607. IEEE, 2020.

\bibitem{albahli2024efficient}
Saleh Albahli.
\newblock Efficient hyperparameter tuning for predicting student performance with bayesian optimization.
\newblock {\em Multimedia tools and applications}, 83(17):52711--52735, 2024.

\bibitem{liu2021projection}
Ling Liu, Bel{\'e}n Mart{\'\i}n-Barrag{\'a}n, and Francisco~J Prieto.
\newblock A projection multi-objective svm method for multi-class classification.
\newblock {\em Computers \& Industrial Engineering}, 158:107425, 2021.

\end{thebibliography}

\end{document}